\documentclass[10pt,twocolumn,letterpaper]{article}

\usepackage{iccv}
\usepackage{times}
\usepackage{epsfig}
\usepackage{graphicx}
\usepackage{amsmath}
\usepackage{amssymb}
\usepackage[accsupp]{axessibility}

\usepackage{booktabs}

\usepackage{float}

\usepackage[pagebackref=true,breaklinks=true,letterpaper=true,colorlinks,bookmarks=false]{hyperref}

\iccvfinalcopy % *** Uncomment this line for the final submission
\ificcvfinal\fi

\def\iccvPaperID{1998} % *** Enter the ICCV Paper ID here

\ificcvfinal\fi

\begin{document}

%%%%%%%%% TITLE
\title{GaFET: Learning Geometry-aware Facial Expression Translation \\ from In-The-Wild Images}

\author{
% All authors must be in the same font size and format.
Tianxiang Ma\textsuperscript{\rm 1,2}\equalcontrib, \ \ 
Bingchuan Li\textsuperscript{\rm 3}\equalcontrib, \ \ 
Qian He\textsuperscript{\rm 3}, \ \ 
Jing Dong\textsuperscript{\rm 2}\thanks{Corresponding author.}, \ \ 
Tieniu Tan\textsuperscript{\rm 2,4}, \\
%Afiliations
\textsuperscript{\rm 1}School of Artificial Intelligence, UCAS \ \ \ 
\textsuperscript{\rm 2}CRIPAC \& NLPR, CASIA\\
\textsuperscript{\rm 3}ByteDance Ltd, Beijing, China \ \ \ 
\textsuperscript{\rm 4}Nanjing University \\
{\tt\small tianxiang.ma@cripac.ia.ac.cn,\{libingchuan,heqian\}@bytedance.com,\{jdong,tnt\}@nlpr.ia.ac.cn}
}

\maketitle
% Remove page # from the first page of camera-ready.
\ificcvfinal\thispagestyle{empty}\fi

% \twocolumn[{

% \maketitle
% \ificcvfinal\pagestyle{empty}\fi

% \vspace{-0.5cm}

% % \renewcommand\twocolumn[1][]{#1}%

% \begin{figure}[H]
% \hsize=\textwidth
% \centering
% \includegraphics[width=2.0\linewidth]{top_show.pdf}
% \caption{High-quality face expression transfer results obtained by our method on the 512 resolution CelebA-HQ and RaFD datasets. The "Cross" represents the cross expression transfer on that two datasets.}
% \label{fig:show}
% \end{figure}
% }]

% \begin{figure*}[t]
% \begin{center}
% \includegraphics[width=1.0\linewidth]{top_show.pdf}
% \end{center}
% \caption{High-quality facial expression translation with GaFET at 512 resolution.}
% \label{fig:show}
% \end{figure*}

%%%%%%%%% ABSTRACT
\begin{abstract}
While current face animation methods can manipulate expressions individually, they suffer from several limitations. The expressions manipulated by some motion-based facial reenactment models are crude. Other ideas modeled with facial action units cannot generalize to arbitrary expressions not covered by annotations. In this paper, we introduce a novel Geometry-aware Facial Expression Translation (GaFET) framework, which is based on parametric 3D facial representations and can stably decoupled expression. Among them, a Multi-level Feature Aligned Transformer is proposed to complement non-geometric facial detail features while addressing the alignment challenge of spatial features. Further, we design a De-expression model based on StyleGAN, in order to reduce the learning difficulty of GaFET in unpaired ``in-the-wild" images. Extensive qualitative and quantitative experiments demonstrate that we achieve higher-quality and more accurate facial expression transfer results compared to state-of-the-art methods, and demonstrate applicability of various poses and complex textures. Besides, videos or annotated training data are omitted, making our method easier to use and generalize.

% While current face animation methods can manipulate expressions individually, they suffer from several limitations. The expressions manipulated by some motion-based facial reenactment models are crude. Other ideas modeled with facial action units cannot generalize to arbitrary expressions not covered by annotations. In this paper, we introduce a novel Unsupervised High-Fidelity Face Expression Transfer (UHFET) framework, which is based on a geometry-aware parametric 3D facial representations and a expression-aware feature aligned Transformer module. Among them, the former focuses on overall geometric-level expression transfer, while the latter deals with detailed expression textures that geometry cannot represent, such as teeth. Further, we design a De-expression model based on pre-trained StyleGAN, in order to reduce the learning difficulty of UHFET's unsupervised training on ``in-the-wild" images. Extensive qualitative and quantitative experiments demonstrate that we achieve higher-quality and more accurate facial expression transfer results compared to state-of-the-art methods, and demonstrate applicability of various poses and complex textures. Besides, videos or annotated training data are omitted, making our method easier to use and generalize.
\end{abstract}

% Geometry-aware Facial Expression Translation (GaFET) framework
% a novel multi-level transformer is proposed to complement non-geometric facial detail features while addressing the alignment challenge of spatial features.

%%%%%%%%% BODY TEXT
\section{Introduction}
Photo-realistic face expression generation and manipulation have rich practical application scenarios, such as portrait photo beautification, face emotion control, and movie industry production. However, expression as a face attribute is highly bound with face shape, texture, lighting and other factors, which makes it difficult to transfer and manipulate effectively. These factors bring a lot of challenges to the expression translation task.

Early GAN-based expression synthesis and translation methods \cite{yeh2016semantic,wang2018facial,song2018geometry,pumarola2018ganimation} suffer several limitations, among which the most painful problem is the low quality of synthesized images. They are usually generated at 128 resolution or below, which naturally loses a lot of expression detail. Another idea \cite{ding2018exprgan,tang2019expression} relies on specific face expression category labels for cross-domain image-to-image translation. The most common are the 8 basic expression categories such as happy, sad, surprised, etc. But these completely limit the diversity of facial expressions. More novel approaches \cite{pumarola2018ganimation,ling2020toward} utilize action units (AUs), which describe the anatomical facial movements that define human expressions in a continuous manifold. It is equivalent to splitting the expression annotations of the whole face image into multiple sets of local expression annotations. Although AUs are more fine-grained than image categories, it is still difficult to accurately describe arbitrary face geometries at such a high level of abstraction.

\begin{figure*}[t]
\begin{center}
\includegraphics[width=0.95\linewidth]{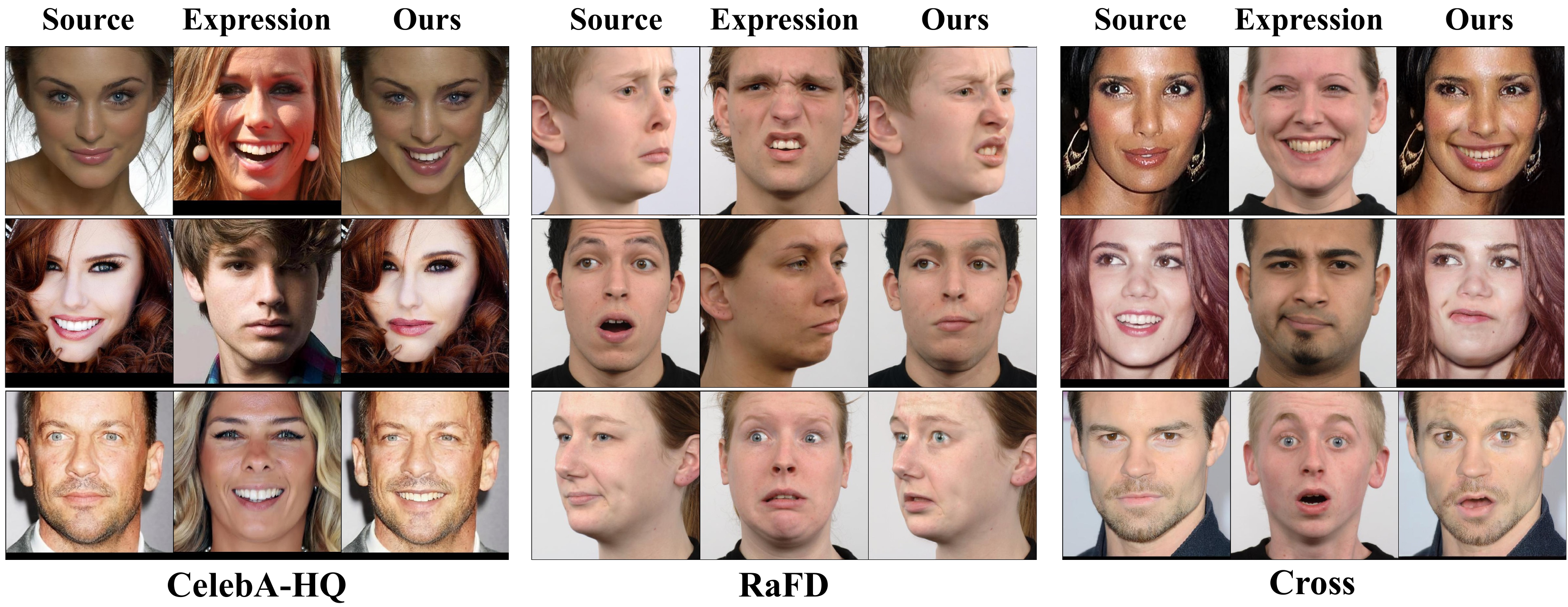}
\end{center}
\caption{High-quality face expression transfer results obtained by our method on the 512 resolution CelebA-HQ and RaFD datasets. The "Cross" represents the cross expression transfer on that two datasets.}
\label{fig:show}
\end{figure*}

Another expression-related face reenactment techniques \cite{siarohin2019first,ha2020marionette,burkov2020neural,wang2021one, doukas2021headgan, drobyshev2022megaportraits} have been widely developed. They usually require a large amount of video data for self-supervised training, and use the reference head motion video to drive the target face during the inference process. However, since large motions and deformations (e.g. pose changes) are the focus of this task, these methods usually ignore fine-grained expression synthesis. Blurring, deformation, and distortion are common in face reenactment methods. In addition, some reenactment methods such as FOMM \cite{siarohin2019first} and DaGAN \cite{hong2022depth} cannot decouple expressions automatically. Therefore, when comparing and evaluating these methods, the reference images of pose and shape pairings are artificially selected to eliminate their disadvantages for expression transfer as much as possible.

% the main causes of bad cases in reenactment tasks, such problems are so difficult that they are negligent in fine expression synthesis.

% Besides, blurring, deformation, and distortion are common in face reenactment methods. 
% Although these methods utilize geometric information, such as 2D landmarks or 3D parametric models, to learn facial animation, they tend to neglect fine expression synthesis due to difficulties such as head pose transfer. 
% For more details, please move to the related works.

Aiming at the difficulties of current expression manipulation, we propose a novel geometry-aware facial expression translation (GaFET) method to improve it. First, we adopt 3D face geometry to represent expressions, in order to avoid class annotations limiting the generative diversity. We introduce state-of-the-art 3D face parametric model EMOCA \cite{danvevcek2022emoca}, which has the most accurate and stable expression representation so far. Specifically, we use the EMOCA encoder to extract expression features from the reference face with other expression-independent features from the source face, and then combine them, and differentiably render a 3D geometric image. Therefore, it has the pose and shape of the source face and maintains the expression characteristics of the reference face.

Moreover, we found that 3D parametric geometry alone cannot fully represent fine expression information. For this reason, we propose a Multi-level Feature Aligned Transformer (MFAT) network that helps to extract texture details from expression image to complement the lack of geometric features, like teeth details. Since the source and expression faces may have different poses and shapes, spatially aligning detailed features and geometric features becomes a challenge. Our proposed MFAT has a global receptive field and can efficiently align local expression features through self-attention and cross-attention mechanisms. We add the multi-level features learned from MFAT to the spatial features learned from 3D geometry in the form of residuals to complete the expression details.

Considering the high acquisition cost of video or labeled images, our model is built on unlabeled and unpaired ``in-the-wild" images, so it is easier to apply more high-quality images for learning. Particularly, we design an emotion removal method based on StyleGAN's latent space editing and incubate a De-expression model to reduce the learning difficulty of unpaired images. It helps GaFET compute the reconstruction loss using the pseudo-paired data constructed by the De-expression model. GaFET is trained in a generative adversarial manner, where we carefully design the generator structure, including proposing a spatial and global modulation module and integrating a multi-level feature deformation module \cite{yang2022text}, while adding an additional local discriminator for quality of detail. Experiments verify the excellent generalization performance of our method. The main contributions can be summarized as follows:

\begin{itemize}
    
    \item We propose a novel geometry-aware facial expression translation framework that achieves high-quality expression transfer.
    
    \item We further propose a Multi-level Feature Aligned Transformer module to extract detailed features and spatially align them with geometric features.
    
    \item The De-expression model is designed based on StyleGAN, which reduces the difficulty of training on unlabeled and unpaired ``in-the-wild" images.
    
    \item Compared with SOTA facial expression manipulation methods, we have clear advantages in performance and ease of use.
    
\end{itemize}

% 先说一下Reenactment的做法，指出它们的缺点，要使用配对的视频数据训练，数据的多样性不足，且缺少显示的表情表示。 然后再说到之前一些单图像表情生成和迁移工作，它们的生成效果一般，分辨率较低，需要提前提取到图像的AU，且方法不够系统。然后开始介绍我们的方法，我们提出一种更系统和鲁棒的无监督单图像表情生成方法，显示的解耦表情特征，和采用先进的生成与特征对齐框架，来实现更精细的表情生成和迁移。

\begin{figure*}[t]
\begin{center}
\includegraphics[width=0.85\linewidth]{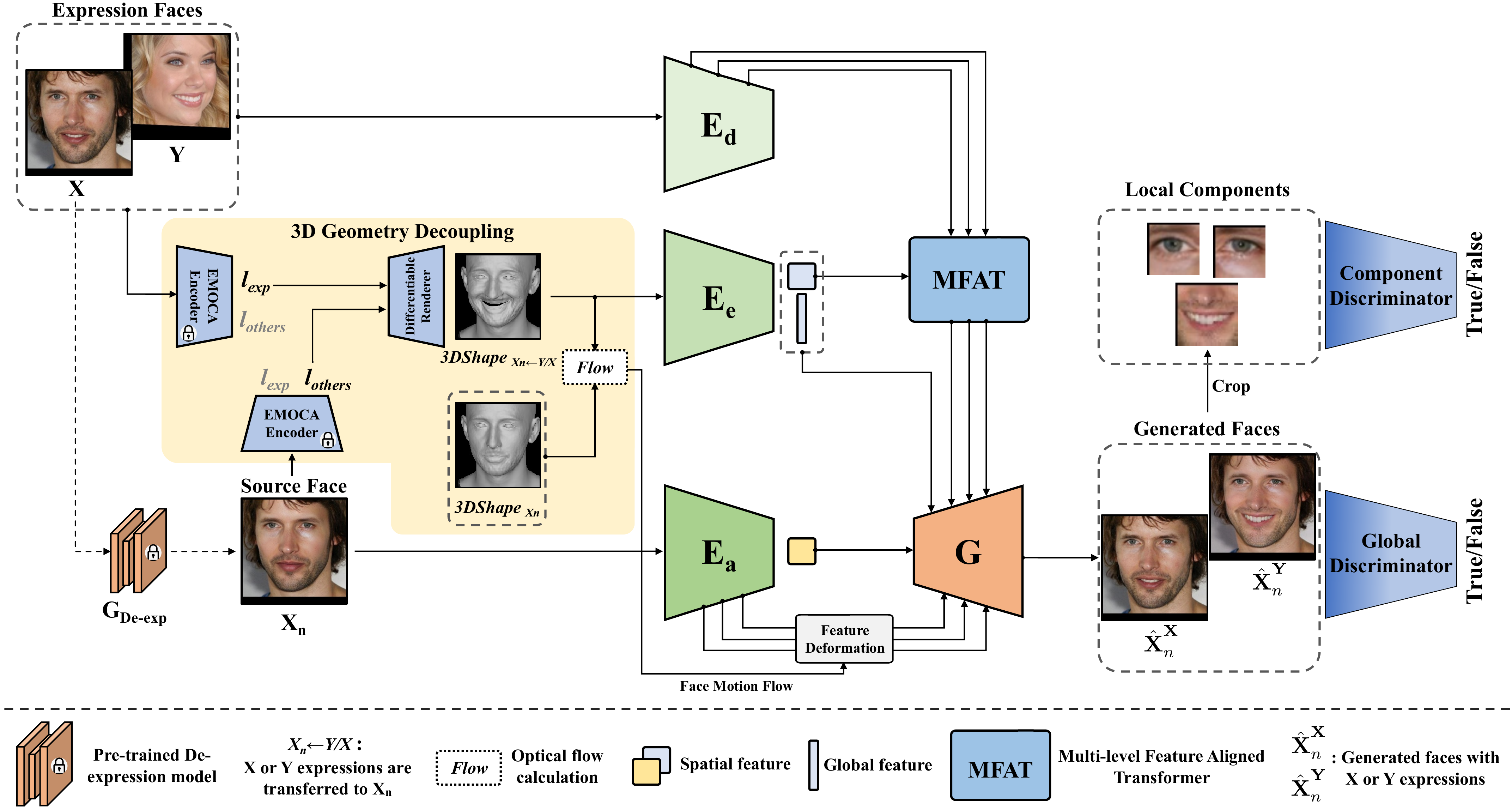}
\end{center}
\caption{The overall framework of our GaFET. The 3D geometry decoupling module provides a geometry-aware expression representation, where expression-related features are derived from reference expression faces, and other expression-independent attributes are derived from source faces. The three encoders $E_d$, $E_e$, and $E_a$ extract features from the expression reference face, 3D shape, and source face, respectively. The MFAT spatially aligns and fuses expression-related features. The pre-trained De-expression model offers pseudo-paired data to help model training. The $\textit{Flow}$ calculates facial geometric motion to deform source face features.}
\label{fig:model}
\end{figure*}

\section{Related Work}

\subsection{Facial Expression Synthesis and Transfer}
Expression synthesis and transfer is a long-studied task. Early approaches \cite{thies2015real, thies2016face2face} relied on professional RGB-D camera to capture the face geometry and target the design of character-specific expression transfer. Later,  Some methods \cite{yeh2016semantic,wang2018facial,ding2018exprgan,song2018geometry} started to explore the optimization of network structure to enhance expression 
transfer.
% Yeh et al. \cite{yeh2016semantic} used autoencoded flow to implement simple face expression editing.
% Wang et al. \cite{wang2018facial} designed a U-Net conditional generative adversarial network to implement face expression generation. Ding et al. \cite{ding2018exprgan} designed an expression controller and used classification network to learn expressive and compact expression code. 
Pumarola et al. \cite{pumarola2018ganimation} proposed to use face action units as conditional input for expression features and added an alpha channel output.
% to let the model focus on learning features at expression-related locations, achieving good results.
% Song et al. \cite{song2018geometry} extracted face key points as geometric guides to generate face expressions. 
Yang and Lim \cite{yang2019unconstrained} used pre-trained StyleGAN to do unconstrained expression transfer.
% they used the W+ layers swapping strategy to transfer expressions and mentioned that layer 4 of the W+ space is sensitive to expressions, but there is little subsequent work to follow up this idea. 
Fan et al. \cite{fan2019controllable} started to consider image to video face expression transfer and extended it with some interesting applications. 
% This method has improved the resolution of expression generation compared to previous methods, but still does not achieve a high quality. 
Wang et al. \cite{wang2019dft} explicitly decoupled face deformations and appearance details by constructing two parallel networks.
Tang et al. \cite{tang2019expression} proposed an expression conditional GAN model and use the binarized expression category vectors to guide model generation and training.
% and they considered both expression transfer and recognition tasks. 
% Yi et al. \cite{yi2020animating} designed a high-quality facial expression animation model using the idea of warping and hyper-resolution, but the generated expression details are not fine enough. Usually, artifacts are introduced in expression transfer,
Wu et al. \cite{wu2020cascade} proposed a progressive network with special attention to improve the expression transfer effect.
% but the quality of generated images does not improve. 
Ling et al. \cite{ling2020toward} also use face action units as conditional input, but they use the variations of AUs to guide expression synthesis.
% and use a network structure in the form of cycle. 
Jourabloo et al. \cite{jourabloo2022robust} implemented 3D reconstruction of facial expressions based on face data captured by VR devices.
% and achieved good results. However, the use of professional equipment limits the applicability of the method. 
% In this paper, we propose a high-fidelity facial expression synthesis method with more systematic model design and advanced technical support, which is able to achieve better performance under qualitative and quantitative evaluation compared to previous methods.

\subsection{Face Reenactment and Talking Head}
Face reenactment and talking head generation are the tasks very relevant to expression synthesis, which aims to transfer the head pose and movement that incorporate expressions. Earlier, Nirkin et al. \cite{nirkin2019fsgan} proposed a recurrent neural network-based face reenactment and face swapping method.
Siarohin et al. \cite{siarohin2019first} first proposed a first-order motion model that is capable of unsupervised learning of face animation.
Tripathy et al. \cite{tripathy2020icface} proposed an interpolable face animator, which is driven by head pose angles and face action units. 
% Yao et al. \cite{yao2020mesh} used the graph convolutional network to learn face motion. 
Gu et al. \cite{gu2020flnet} used landmark as the driving input for talking face.
% and used a face image bank as source inputs to make the local features of the generated face, such as eyes and teeth, more refined. 
Ha et al. \cite{ha2020marionette} proposed a series of modules to achieve high-quality face reenactment. 
Burkov et al. \cite{burkov2020neural} designed a neural head reenactment model that drives the head through a latent pose representation.
% while allowing automatic segmentation of foreground. The method demonstrates the neural network latent space decoupling capability in face reenactment. 
Zhang et al. \cite{zhang2020freenet} used face landmark as the vehicle to train a unified landmark converter and a geometry-aware generator.
% Huang et al. \cite{huang2020learning} used video data to train an identity-independent latent space.
% also using both landmark space and image space features to learn face reenactment. 
Wang et al. \cite{wang2021one} proposed a neural talking-head generation method based on decoupled keypoint representation.
Recently, Hong et al. \cite{hong2022depth} introduced a self-supervised face-depth learning method to assist in talking-head generation.
Yin et al. \cite{yin2022styleheat} used a pre-trained StyleGAN model and learned spatial transformation in the latent space to achieve talking face generation.
% Yang et al. \cite{yang2022text} introduced a real-time high-resolution face reenactment method based on 3DMM parametric model.
Doukas et al. \cite{doukas2021headgan} introduced high-resolution face reenactment method based on 3DMM parametric model and Yang et al. \cite{yang2022text} further improved it into a real-time algorithm. Yang et al. \cite{yang2023designing} proposed a 3d-aware face editing method, which can drive facial expressions. Lyu et al. \cite{lyu2023deltaedit} introduced a text-driven face manipulation approach, which is suitable for simple expression editing. Hsu et al. \cite{hsu2022dual} proposed a dual-generator structure to learn face reenactment in large poses. Drobyshev et al. \cite{drobyshev2022megaportraits} designed specific architecture and training procedure that encourages disentanglement between motion and appearance.
% In this paper, we also compare some face reenactment and talking head methods under appropriate experimental setting, and our approach outperforms SOTA methods for expression transfer.
% for large pose and complex style face images. 

\begin{figure}[t]
\begin{center}
\includegraphics[width=1.0\linewidth]{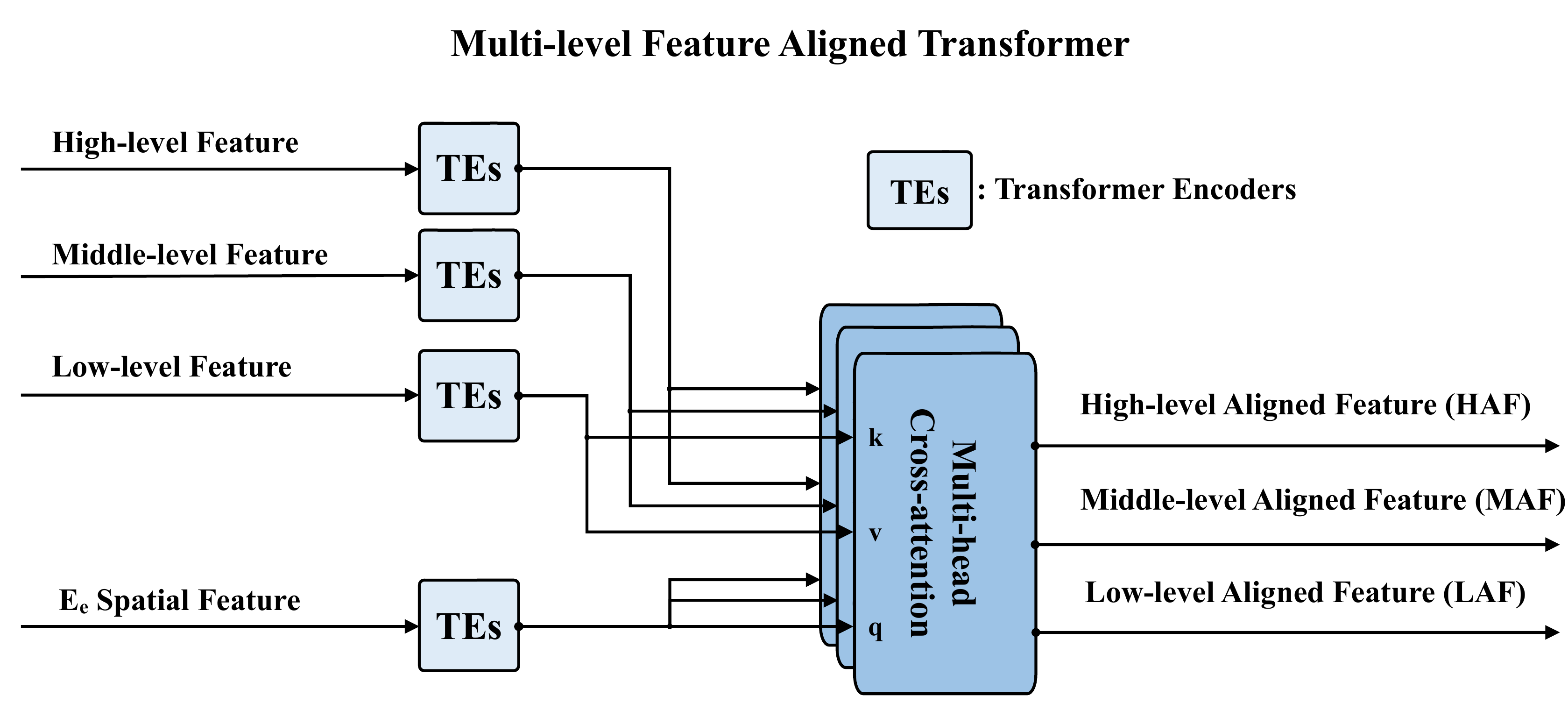}
\end{center}
\caption{Detailed structure of the MFAT module.}
\label{fig:MFAT}
\end{figure}

\begin{figure}[t]
\begin{center}
\includegraphics[width=1.0\linewidth]{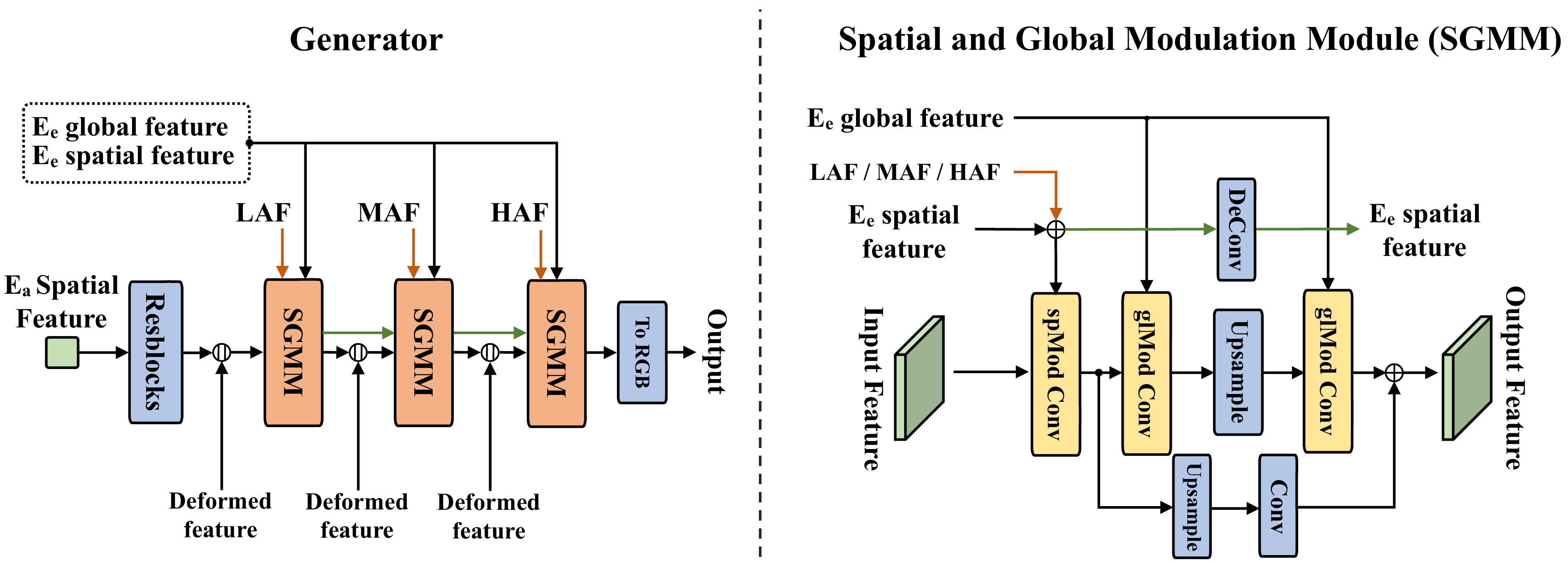}
\end{center}
\caption{Detailed structure of the facial expression Generator.}
\label{fig:Gen}
\end{figure}

\section{Approach}
In this section, we will introduce the geometry-aware facial expression translation framework in detail. Since there is no paired or annotated facial expression data, we designed a semi-supervised approach to help model training by introducing a portion of pseudo-paired data. Consider a source face $X$ and a reference expression face $Y$, and expect to generate a new face that maintains the identity of $X$ and has the expression of $Y$, denoted by $\hat{X}^Y$. To construct pseudo-paired data, we introduce a pre-trained De-expression model $G_\text{De-exp}$, which removes as much expression as possible from a source face $X$ to generate neutral face $X_n$. Therefore, we can use $X_n$ and $X$ as the source image and reference image respectively, generate $\hat{X}_{n}^X$, and perform reconstruction constraints with $X$. Meanwhile, we also use the unpaired $Y$ as a reference expression to generate $\hat{X}_{n}^Y$, and focus more on the identity-independent facial geometry loss. The framework optimizes these two input modes in parallel for sufficient learning. The overall framework is shown in the Figure \ref{fig:model}. In the following parts, the model details are presented.

\subsection{3D Geometry Decoupling}
Since facial expression features and the facial shape are highly correlated, it is desirable to effectively separate the two attributes. We introduce an emotion-driven face capture method EMOCA \cite{danvevcek2022emoca}, which can well decouple the expression, shape, and pose attributes of the face. We utilize EMOCA encoder to extract expression-independent feature coefficients ($l_{\text{others}}$) such as global pose and shape from the source face, and expression and jaw feature coefficients ($l_{\text{exp}}$) from the reference expression face. These coefficients are combined and the 3D geometry is decoded and reconstructed based on FLAME \cite{li2017learning}. Then we utilize the expression encoder $E_e$ to learn spatial and global features from input 3D geometry. Besides, we also render the motion flow between the source 3D vertices and the 3D vertices of  combination coefficients to facilitate feature warping operations in the generator. The modeling process for this part is shown on the left side of the figure \ref{fig:model}.

\subsection{Multi-level Feature Aligned Transformer}
Since the 3D geometry lacks the detailed information of the face texture, like teeth, it cannot fully represent the expression. The straightforward idea is to extract detailed texture features from reference expression face to complement the missing information in 3D geometry. Since the poses and shapes of the source face and the reference face are inconsistent, it is unreasonable to directly overlay the texture features with the geometric features. To address this issue, we propose a novel multi-level feature aligned transformer (MFAT) that can spatially fuse and align the two features.

The structure of the module is shown in the Figure \ref{fig:MFAT}. First, The encoder $E_d$ extracts multi-level features of the reference expression face, divided into high, middle and low level features. The spatial feature learned by $E_e$ are simultaneously obtained. Then, they are passed through a set of transformer \cite{vaswani2017attention} encoders separately to learn the self-attention features. This process can adaptively learn useful information of features with the help of a global multi-head self-attention mechanism, and is not affected by the spatial alignment problem. Finally, for each level, we use a multi-head cross-attention module that utilizes the spatial features extracted by $E_e$ to query the matched features on different level, and outputs multi-level aligned features. After learning, this module can supplement the missing details of geometry.

\subsection{Face Expression Generation}
For a given source face, we use the appearance encoder $E_a$ to extract spatial features as input to the expression generator backbone. The detailed structure of the proposed expression generator is shown in the Figure \ref{fig:Gen}. Where SGMM is used to fuse the features extracted from different modules. Within each SGMM, the spatial and global features extracted by $E_e$ work on different modulation networks, called spatial modulation network and global modulation network, respectively. They are similar in form to the (de)modulation modules in StyleGAN \cite{karras2020analyzing}, and the difference lies in whether the input modulation features carry a spatial dimension. Using the global features can help to better guide the synthesis process, and its effectiveness has been demonstrated in the ablation study. We add multi-level expression-aware aligned features learned from MFAT to the $E_e$ spatial feature as residuals. This approach can effectively combine both features to get our complete expression representations. The new spatial feature obtained at each level is passed through the deconvolutional layer to the next level of the network. 

We employ a U-Net-like network structure to obtain features from $E_a$ at different scales and inject them into the generator. Meanwhile, the feature deformation \cite{yang2022text} network is used to morph the input multi-scale $E_a$ features by the motion flow computed from two 3D vertices. This makes it easier for the generator to perceive geometric changes in expressions. Finally, the last output layer of the generator converts the features into an RGB image, which is an expression-translated face.

\begin{figure}[t]
\begin{center}
\includegraphics[width=1.0\linewidth]{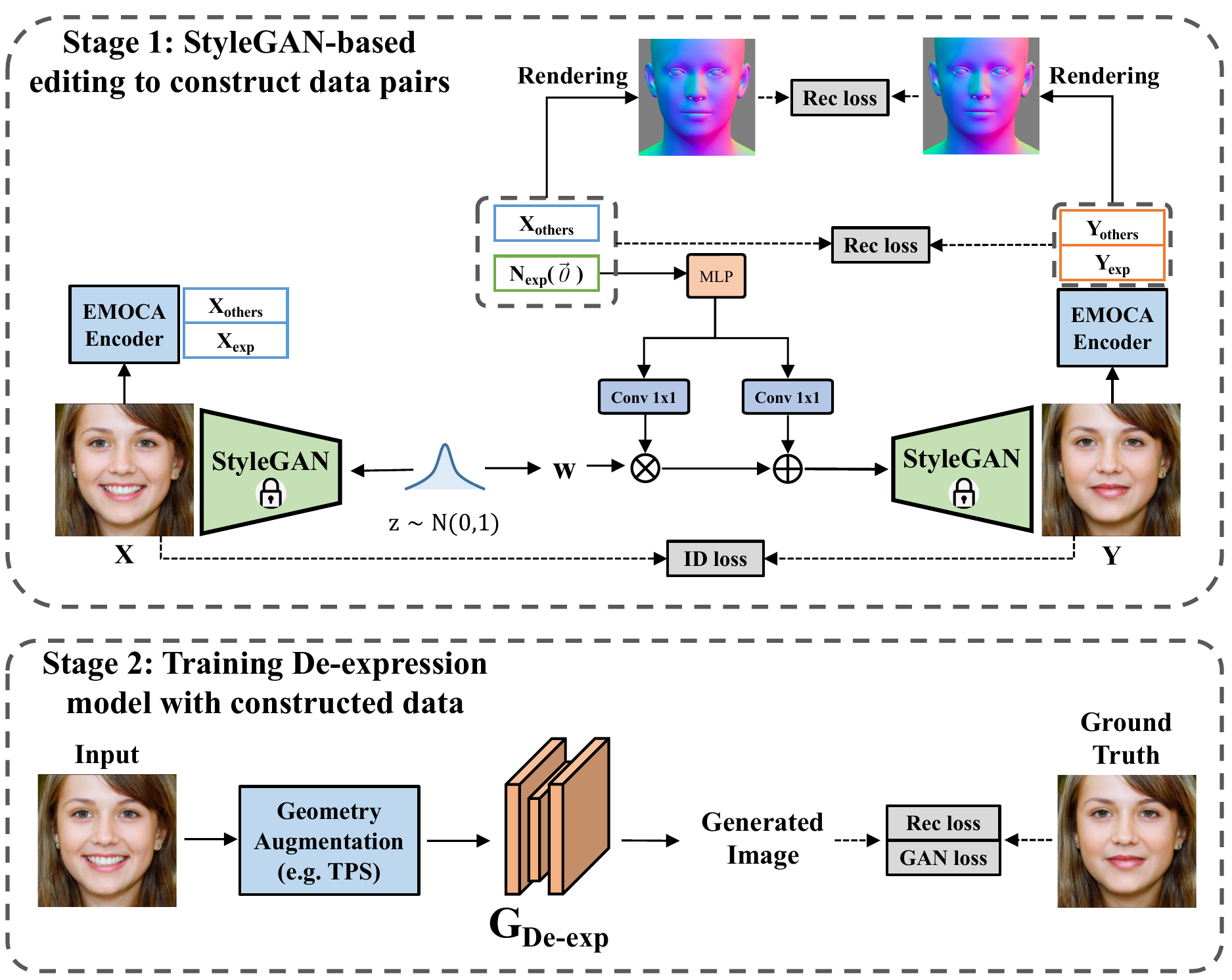}
\end{center}
\caption{De-expression model training process. }
\label{fig:De-exp}
\end{figure}

\subsection{Pre-trained De-expression Model}
The pre-trained De-expression model is an important part of our GaFET framework to implement pseudo-paired data training. The training details of the De-expression model are shown in Figure \ref{fig:De-exp}. The main idea is to train a StyleGAN-based editing model to construct paired expression faces so that a De-expression model can be trained. First, we utilized a pre-trained StyleGAN \cite{karras2020analyzing} to train a StyleGAN-based editing model as in the upper part of Figure \ref{fig:De-exp}. Specifically, We use the EMOCA to extract expression coefficients and other expression-unrelated coefficients from the face generated by random sampling. The neutral expression code, i.e., zero vector, is used as the modulation parameter in editing the $w$ code to guide the generation of the corresponding De-expression faces. The training of this editing model (MLP and Convolutional networks) is supervised by the reconstruction loss of 3DMM coefficients and rendered normal maps, while using identity loss to constraint the identity of edited face. Second, we use the trained StyleGAN-based editing model to generate 50,000 paired faces (the face and its neutral expression face), and learn a De-expression model with U2Net \cite{qin2020u2} structure as shown in the lower part of Figure \ref{fig:De-exp}. Further, geometric augmentations (e.g. Thin Plate Spline) are used to improve the robustness of the De-expression model during training. Please refer to the supplementary material for the De-expression effects of this model.

\subsection{Training}
\noindent \textbf{Discriminators.} We employ two discriminators to assist in the training of our expression translation model. A global discriminator is used to discriminate the overall image, and a component discriminator is used to discriminate the eye and mouth regions of the face. Since most of the expression variations are in the mouth and eye regions, the introduction of component discriminator can help our model to better generate local expression details. During training, we extract fixed size patches of the eye and mouth regions using a pre-trained face landmark detector, and feed them into the component discriminator, which has the same structure as the global discriminator except that the number of model parameters is smaller. 

\noindent \textbf{Reconstruction Loss.} In the training stage, we alternately feed pseudo-paired faces $(X_{n},X)$ and unpaired faces $(X,Y)$ as model inputs. When the inputs are pseudo-paired $(X_{n},X)$, we can use the reconstruction loss to train the model with the following equation,
\begin{equation}
\mathcal{L}^{\text{pseudo}}_{\text {rec}}=\left\|X - \hat{X}_{n}^X\right\|_{1} + \lambda \cdot lpips(X, \hat{X}_{n}^X),
\end{equation}
where the first part is the $L_1$ norm reconstruction loss, and the second part is the reconstruction loss calculated using Learned Perceptual Image Patch Similarity (lpips) \cite{zhang2018unreasonable}. $\lambda$ is the weight to balance loss terms. 

Except for the reconstruction loss, the rest of the losses we used are the same for different input conditions. In the following, we present each loss by using $(X_n,Y)$ as inputs.

\begin{figure*}[t]
\begin{center}
\includegraphics[width=0.8\linewidth]{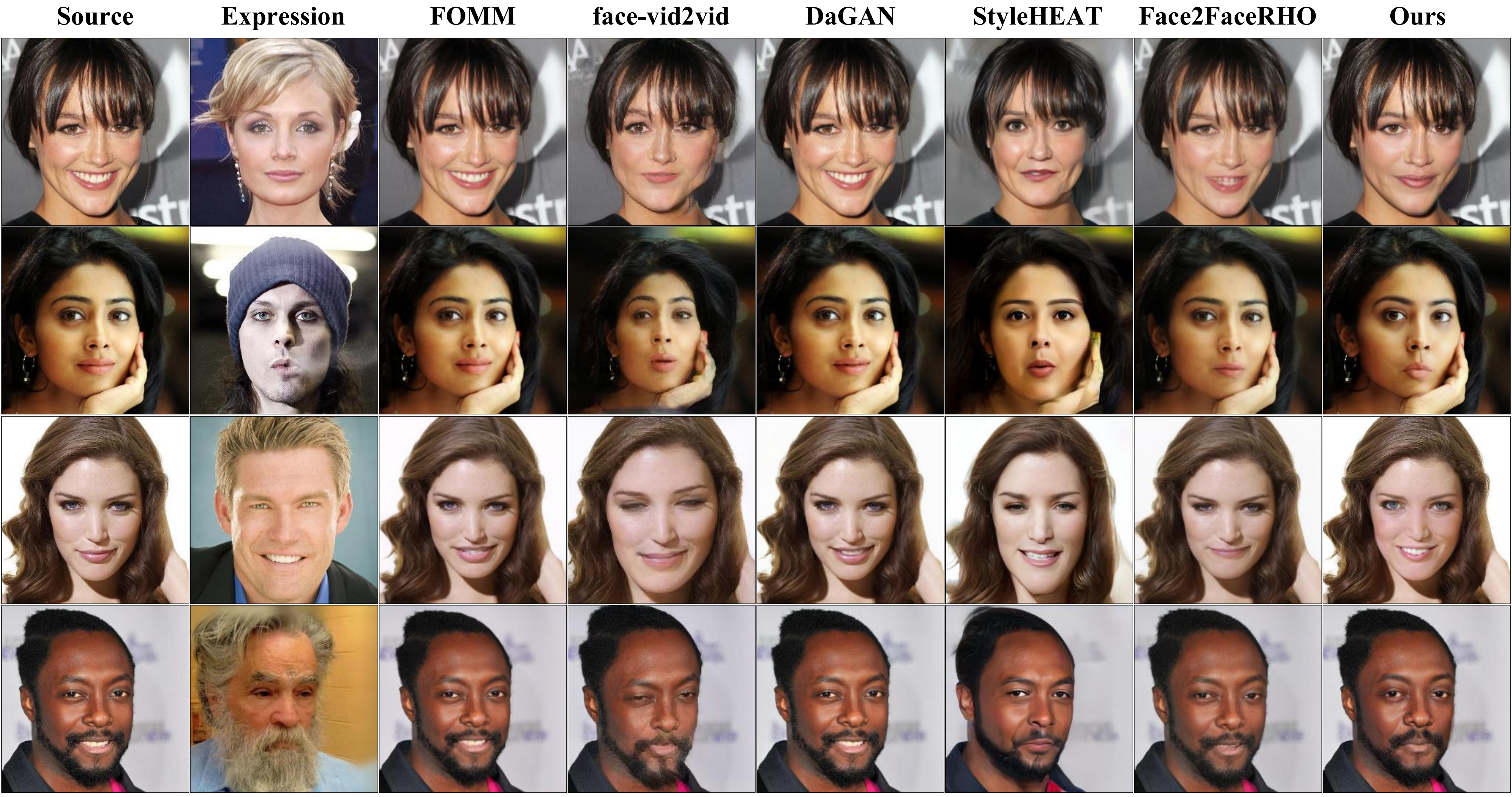}
\end{center}
\caption{Qualitative comparisons with face reenactment methods on expression translation.}
\label{fig:compare_with_reenact1}
\end{figure*}

\begin{figure*}[t]
\begin{center}
\begin{minipage}[t]{0.49\textwidth}
\begin{center}
\includegraphics[width=1.0\linewidth]{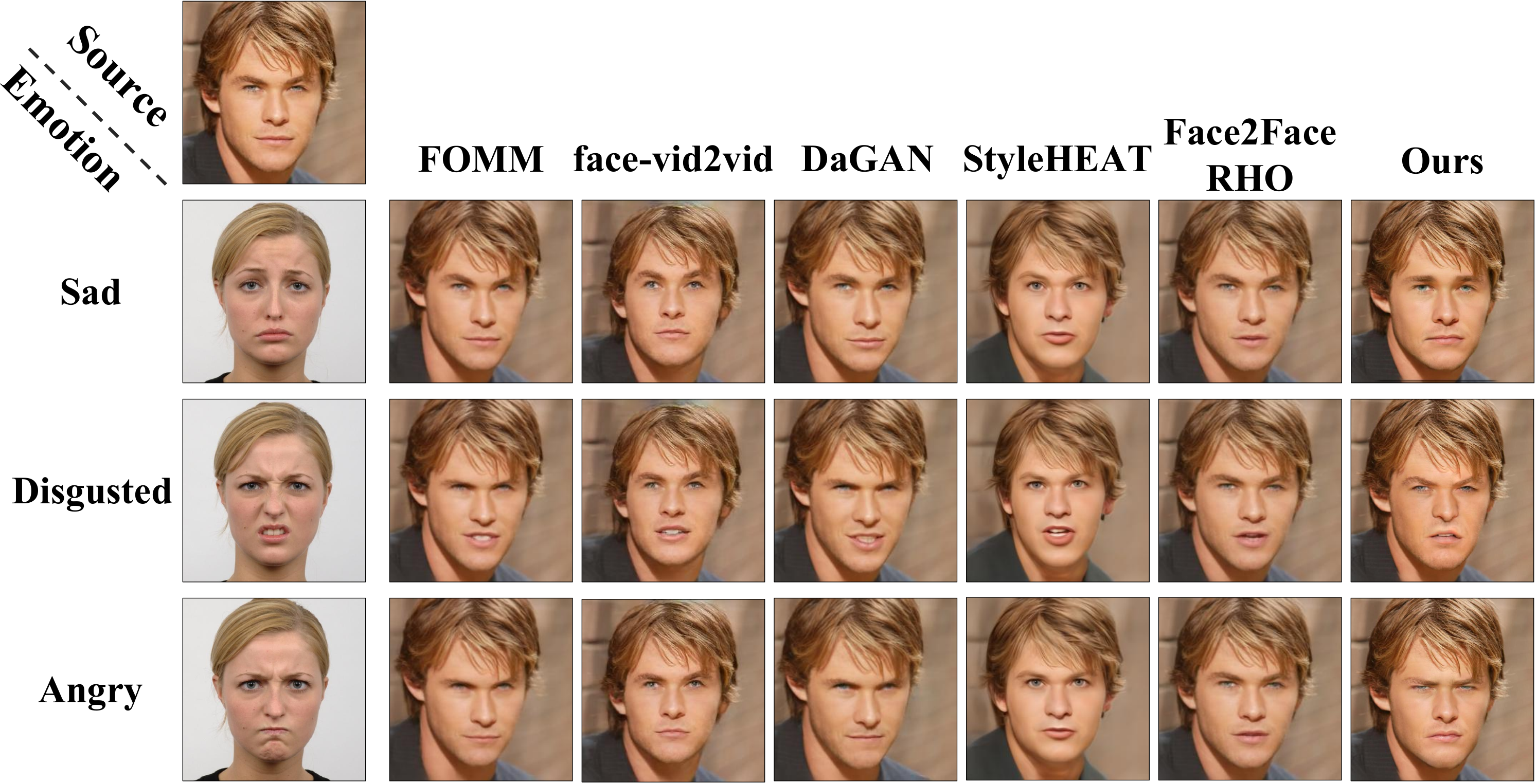}
\end{center}
\end{minipage}
\begin{minipage}[t]{0.49\textwidth}
\begin{center}
\includegraphics[width=1.0\linewidth]{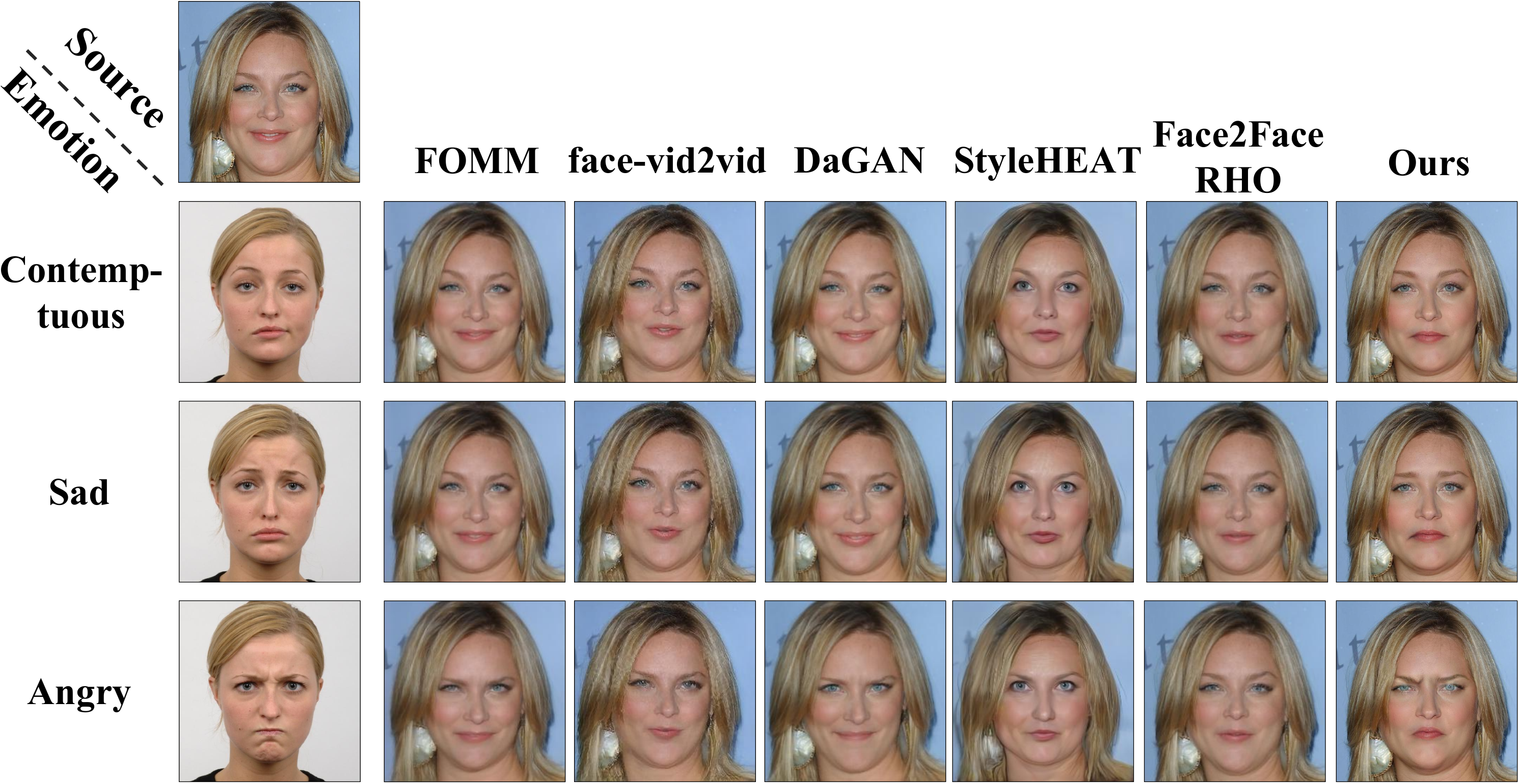}
\end{center}
\end{minipage}
\end{center}
\caption{Qualitative comparisons with face reenactment methods on emotion manipulation.}
\label{fig:compare_with_reenact2}
\end{figure*}

\noindent \textbf{Expression Losses.} For our training task the core purpose is to transfer Y's expression to X. We take two approaches to constrain the expression attribute of the generated face to be consistent with that of Y. First we extract the expression code of the generated face and other expression-independent codes using the same method as in the 3D geometry decoupling module. Then we replace the expression-independent codes with the Y counterpart, and use a differentiable renderer to obtain the 3D shape. We compute the reconstruction loss between that 3D shape and the reconstructed 3D shape of Y as the formula below,
\begin{equation}
\mathcal{L}_{\text {exp}}^{\text{shape}}=\left\|S(l_{\text{exp}}^{\hat{X}}, l_{\text{others}}^Y) - S(l_{\text{exp}}^{Y}, l_{\text{others}}^Y) \right\|_{1},
\end{equation}
where $S(\cdot,\cdot)$ represents the 3D shape rendering operation, $l_{\text{exp}}^{\hat{X}}$ represents the expression feature code extracted from $\hat{X}$, and the others are the similar. This approach achieves expression consistency constraint based on 3D reconstruction, and is superior to direct expression code constraint. 

Earlier we mentioned that the 3D geometry expression representation lacks textured details. So we introduce an additional facial action unit loss. We use a SOTA face action unit detection model MLCR \cite{niu2019multi}, which can detect 12 action units in the face, covering cheek and eye details. We apply this model to extract action units in generated face $\hat{X}_{n}^Y$ and expression face $Y$, respectively, and then calculate the similarity loss as the following formula,
\begin{equation}
\mathcal{L}_{\text {exp}}^{\text{AU}}= 1 - cos\_sim(AU(\hat{X}_{n}^Y), AU(Y)),
% \frac{AU(\hat{X}_{n}^Y) \cdot AU(Y)}{  \|  AU(\hat{X}_{n}^Y) \| \cdot \left\| AU(Y) \right\| },
\end{equation}
where $AU(\cdot)$ represents the AU feature embedding obtained using the face action unit detector. $cos\_sim(\cdot,\cdot)$ means cosine similarity.  

% Additionally, we also extract eye region landmarks using the pre-trained face landmark detector and calculate the opening rate for the left and right eyelids separately. The calculation formula of the eyelid opening rate and the corresponding loss function are as follows,
% \begin{equation}
% \begin{aligned}
% &\mathcal{L}_{\text {exp}}^{\text{eyelid}} = \left\|\mathcal{R}(\hat{X}_{n}^Y) - \mathcal{R}(Y)\right\|_{2}, \\
% &\mathcal{R}(x) = \frac{d(k^{up}(x), k^{down}(x))}{d(k^{left}(x), k^{right}(x)) + \epsilon},
% \end{aligned}
% \end{equation}
% where $\mathcal{R}(x)$ represents the eyelid opening rate calculation method, $d$ is the distance function, and the four $k$ represent the coordinates of the four landmarks at the upper, lower, left, and right of the eyelid region, respectively. The $\epsilon$ is a small value to avoid division by zero.

\noindent \textbf{Contextual Loss.} In order to make the style and texture of the generated face consistent with the source face, we use Contextual Loss \cite{mechrez2018contextual} to constrain the appearance features of the generated face with the following formula,
\begin{equation}
\mathcal{L}_{\text{cx}}=\sum_{i} \omega_{i}[-\log (C X(\phi_{i}(\hat{X}_{n}^Y), \phi_{i}(X_{n})))],
\end{equation}
where $CX(\cdot,\cdot)$ stands for contextual similarity, $\phi_{i}$ is the $i$th layer of low-level features in the pre-trained VGG-19 network, which contain mainly the rich stylized texture of the image, and $\omega_{i}$ is the weight coefficients of the different layers.

% \noindent \textbf{Identity Loss.} We use a pre-trained face recognition network ArcFace \cite{deng2019arcface} to extract the identity vectors of the generated face and the source face, and compute the cosine similarity loss,
% \begin{equation}
% \mathcal{L}_{\text{id}}= 1 - cos\_sim(Id(\hat{X}_{n}^Y), Id(Y)),
% % \frac{Id(\hat{X}_{n}^Y) \cdot Id(Y)}{  \|  Id(\hat{X}_{n}^Y) \| \cdot \left\| Id(Y) \right\| },
% \end{equation}
% where $Id(\cdot)$ denotes the face identity embedding obtained using the face recognition network. 

\noindent \textbf{Adversarial Losses.} During training, we apply two sets of adversarial losses for the global discriminator and the component discriminator, and adapt an R1 regularization loss term for each discriminator. The two groups of adversarial losses are denoted as $\mathcal{L}_{\text{adv}}^{\text{global}}$ and $\mathcal{L}_{\text{adv}}^{\text{comp}}$.

In summary, we optimize the overall objective as follows,
\begin{equation}
\begin{aligned}
\mathcal{L}_{\text{total}} &= \lambda_{\text{rec}} \mathcal{L}^{\text{pseudo}}_{\text {rec }}+\lambda_{\text{exp}}^{\text{shape}} \mathcal{L}_{\text {exp}}^{\text{shape}} + \lambda_{\text{exp}}^{\text{AU}} \mathcal{L}_{\text{exp}}^{\text{AU}} + \lambda_{\text{cx}} \mathcal{L}_{\text {cx}} \\ & + \lambda_{\text{adv}}^{\text{global}} \mathcal{L}_{\text{adv }}^{\text{global}} + \lambda_{\text{adv}}^{\text{comp}} \mathcal{L}_{\text{adv}}^{\text{comp}}.
\end{aligned}
\end{equation}

The weights of the overall objective are $\lambda_{\text{exp}}^{\text{shape}}$=150, $\lambda_{\text{exp}}^{\text{AU}}$=50, and the rest are 1.0. We set these loss weights to equalize the different loss items.

\section{Experiments}
We conduct sufficient qualitative and quantitative experiments to compare SOTA methods, and demonstrate the superior performance of our method.
\subsection{Implementation Details}

\noindent \textbf{Datasets.} Since our method can be trained using unlabeled and unpaired ``in-the-wild" data, we use the FFHQ \cite{karras2019style} dataset and AffectNet \cite{mollahosseini2017affectnet} dataset as the training sets. FFHQ contains 70,000 high-quality images with rich identity attributes and style textures. AffectNet is a large-scale expression face dataset with in-the-wild images. We combined these two datasets to construct a training set of 100,000 images, and extracted face images using the same alignment as EMOCA \cite{danvevcek2022emoca}. Then we train 256 and 512 resolution version models for evaluation and comparison. In addition, to validate ``in-the-wild" expression transfer effect, we use the CelebAHQ \cite{karras2017progressive} and RaFD \cite{langner2010presentation} datasets as the test sets, and other compared methods are similarly tested on both datasets.

\begin{figure}[t]
\begin{center}
\includegraphics[width=1.0\linewidth]{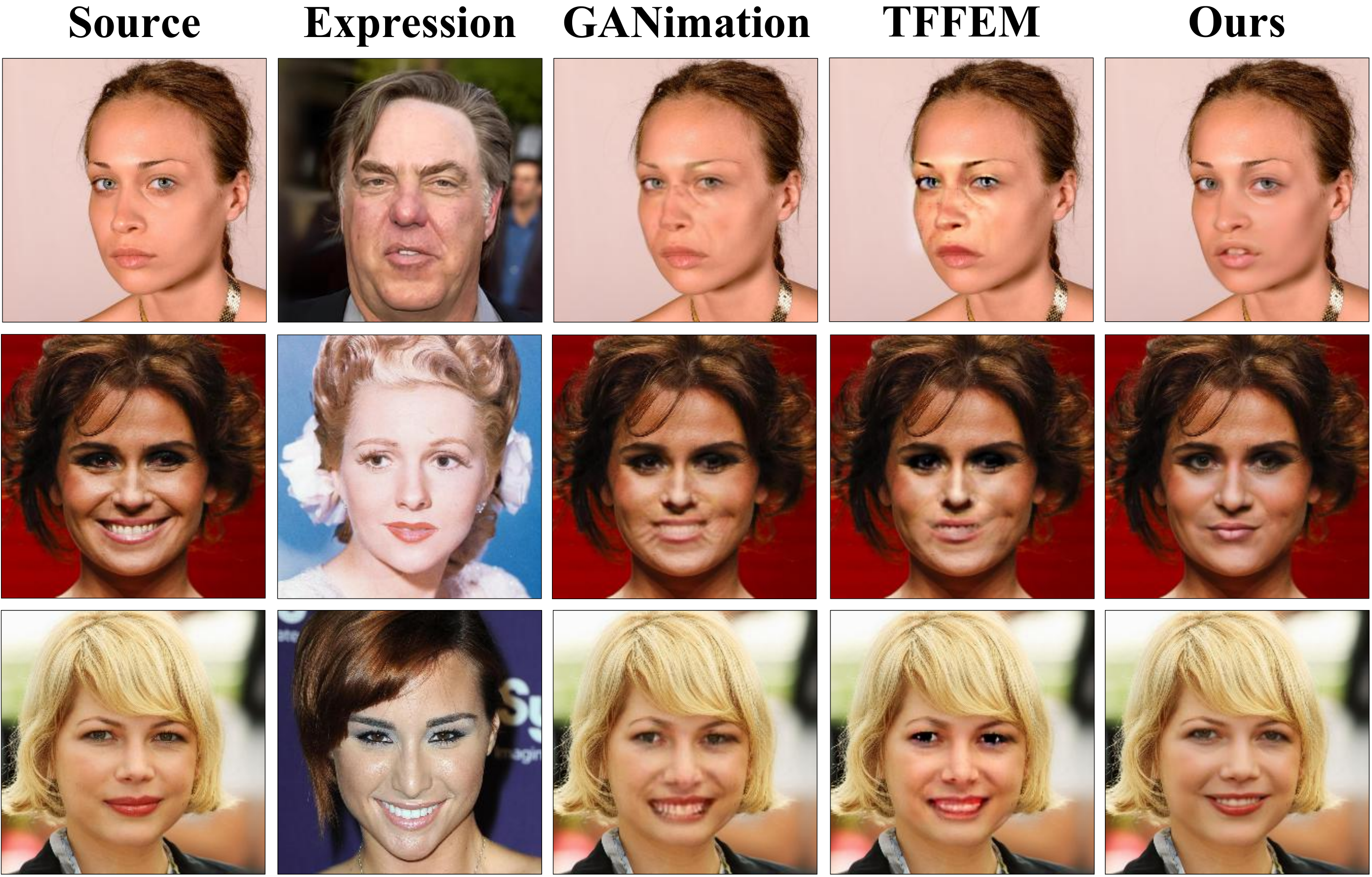}
\end{center}
\caption{Qualitative comparisons with facial expression transfer methods.}
\label{fig:compare_trans}
\end{figure}

\begin{table}[t]
\begin{center}
\setlength{\tabcolsep}{1.3mm}{
\begin{tabular}{l|c|c|c|c}
% \hline
\toprule
% \noalign{\smallskip}
& {FID $\downarrow$} & {CSIM $\uparrow$} & {AED $\downarrow$} & {AU-H $\downarrow$}\\
\hline
% \noalign{\smallskip}
{FOMM \cite{siarohin2019first}}         & {17.8} & {0.681} & {0.289} & {0.267} \\
{Facevid2vid \cite{wang2021one}}        & {42.1} & {0.557} & {0.269 } & {0.287} \\
{DaGAN \cite{hong2022depth}}            & {15.6} & {0.751} & {0.274 } & {0.225 } \\
{StyleHEAT \cite{yin2022styleheat}}     & {35.8} & {0.623} & {0.265 } & {0.195} \\
{Face2FaceRHO \cite{yang2022text}}      & {17.1} & {0.748} & {0.284 } & {0.218} \\
{Ours}                                  & \textbf{11.2}  & \textbf{0.769} & \textbf{0.258} & \textbf{0.174} \\
% \hline
\bottomrule
\end{tabular}}
\end{center}
\caption{Quantitative evaluation of our method and face reenactment methods.}
\label{table:metrics1}
\end{table}

\noindent \textbf{Baselines.} First, we compare SOTA expression translation methods GANimation \cite{pumarola2018ganimation} and TFFEM \cite{ling2020toward}. Second, considering expression translation has been gradually integrated into face reenactment and talking head, we also compare several SOTA face reenactment and talking head methods, FOMM \cite{siarohin2019first}, Facevid2vid \cite{wang2021one}, DaGAN \cite{hong2022depth}, StyleHEAT \cite{yin2022styleheat} and Face2FaceRHO \cite{yang2022text}. Besides, to make a fair comparison with them, we choose the source and reference face with similar head pose, and focus on evaluating expression translation accuracy.
% Since some reenactment methods cannot control the expression alone without affecting the head pose, we try to assign the source face and the reference face to test in the same pose. 

% Please refer to the evaluation section for detailed settings.

% \noindent \textbf{Metrics.} We refer to the evaluation metrics commonly used in expression transfer methods and face reenactment methods to reasonably evaluate our method with baselines. Frechet Inception Distance (FID) is used to measure the realism of the generated images. Cosine Similarity (CSIM) is used to evaluate the identity attribute gap. Average Expression Distance (AED) and Action Units Hamming distance (AU-H) 

\noindent \textbf{Experiment Settings.} We train our model using the Adam optimizer with $\beta_1=0.0$ and $\beta_2=0.99$, and the initial learning rate is 0.002. Please refer to supplementary materials for more training details.

\begin{figure}[t]
\begin{center}
\includegraphics[width=1.0\linewidth]{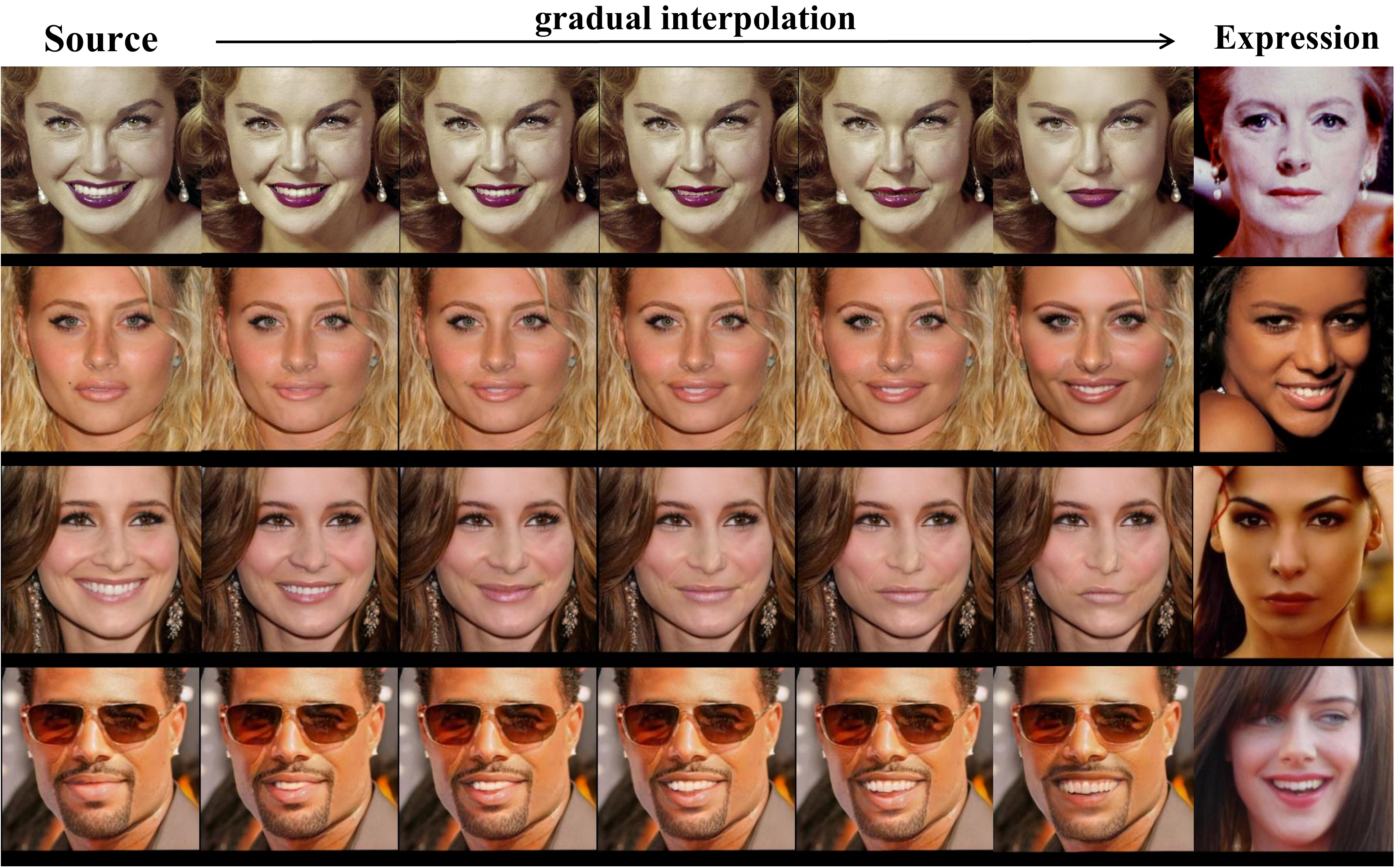}
\end{center}
\caption{Progressive manipulation of expression translation.}
\label{fig:gradual}
\end{figure}

\begin{table}[t]
\begin{center}
\setlength{\tabcolsep}{4.0mm}{
\begin{tabular}{l|c|c|c}
% \hline
\toprule
% \noalign{\smallskip}
& {IS $\uparrow$} & {ACD $\downarrow$} & {ED $\downarrow$}\\

\hline
% \noalign{\smallskip}
{GANimation \cite{pumarola2018ganimation}}  & {2.71} & {0.396} & {0.315}  \\
{TFFEM \cite{ling2020toward}}         & {2.74} & {0.379} & {0.273} \\
{Ours}          & \textbf{2.84} & \textbf{0.350} & \textbf{0.266}  \\
% \hline
\bottomrule
\end{tabular}}
\end{center}
\caption{Quantitative evaluation of our method with facial expression transfer methods.}
\label{table:metrics2}
\end{table}

\subsection{Quantitative and Qualitative Evaluation}

We first compare with SOTA open-source available face reenactment and talking head methods, focusing on evaluating image quality and expression accuracy. We use FID \cite{heusel2017gans} to measure image realism and Cosine Similarity (CSIM) of identity embeddings extracted from CurricularFace \cite{huang2020curricularface} to measure identity preservation after expression manipulation. Average Expression Distance (AED) \cite{ren2021pirenderer} and Action Unit Hamming Distance (AU-H) \cite{baltruvsaitis2015cross} are used to count the accuracy of expression translation. As shown in Table \ref{table:metrics1}, our method achieves state-of-the-art results in each measurement. Figure \ref{fig:compare_with_reenact1} and \ref{fig:compare_with_reenact2} highlight some results. Among them, since FOMM and DaGAN cannot decouple expressions automatically, we artificially separate expressions using two reference images paired with pose and shape during inference. It can be seen that the driven source image completely maintains the pose and shape, but the expression changes are still insufficient. Facevid2vid is the decoupling of 3D pose, translation and expression inside the model. However, when we only modify the expression coefficients, the face undergoes an unsatisfactory drastic change. StyleHEAT generates high-quality maps, but significantly changes identity and pose. The pose of Face2FaceRHO is well maintained but the expression changes are weak. In contrast our method can transfer expressions more accurately while keeping the source face identity and appearance unchanged.
% All the methods compared above suffer from common appearance problems, such as distortion from facial deformation or blurred teeth.

We also compare quantitatively and qualitatively with SOTA expression transfer methods. We follow the evaluation metrics of the TFFEM for quantitative evaluation. Inception Score (IS) \cite{salimans2016improved} is used to evaluate the image quality. Average Content Distance (ACD) is used to measure the identity attribute gap by face recognition network. Expression Distance (ED) calculates the $l_2$-distance between the AU values of the generated face and the reference face. Table \ref{table:metrics2} shows the comparison results, where our method achieves higher scores. Figure \ref{fig:compare_trans} shows that our method achieves better results in terms of both image quality and expressive detail.

\begin{figure}[t]
\begin{center}
\includegraphics[width=1.0\linewidth]{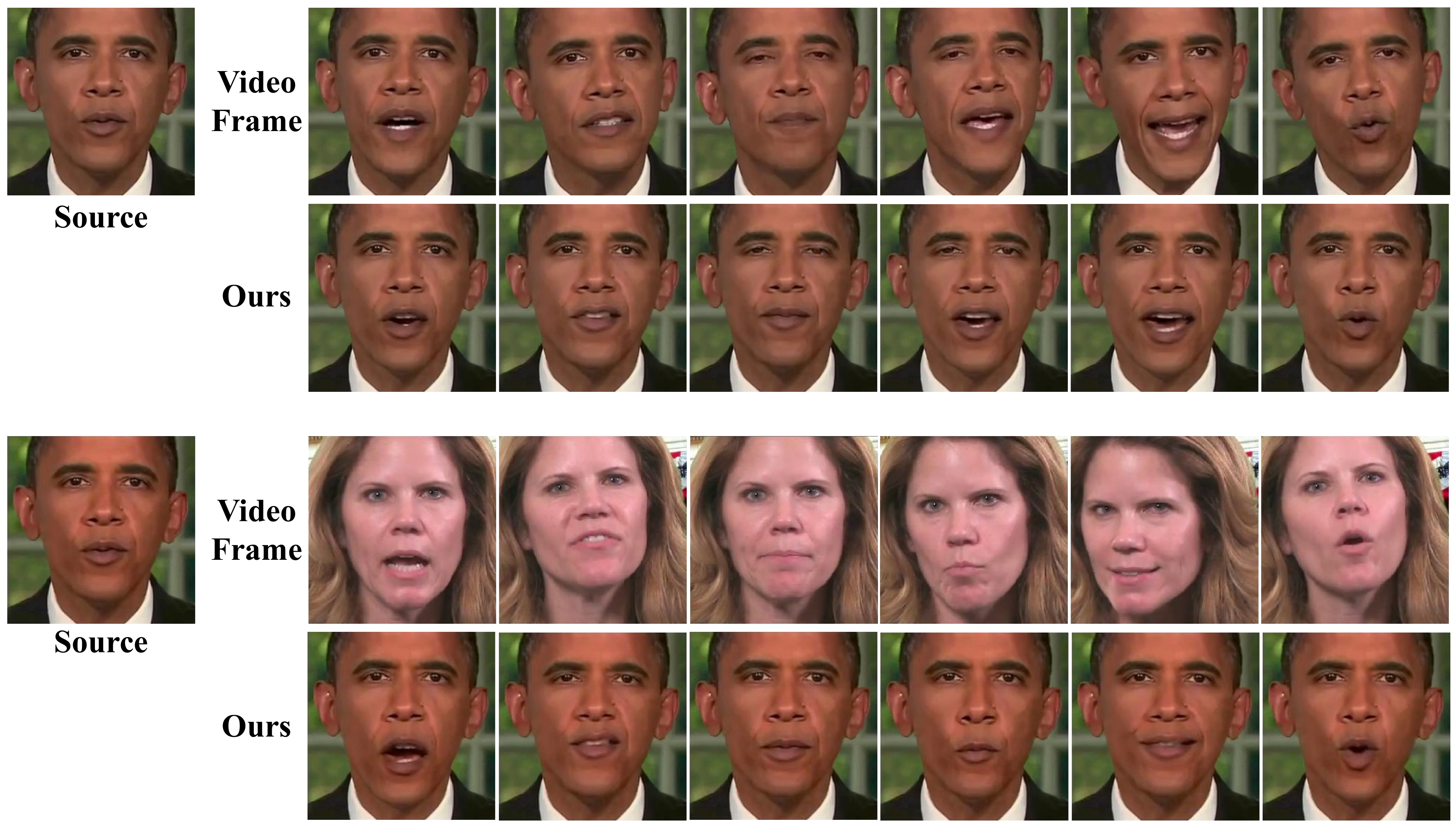}
\end{center}
\caption{The expression translation results on the video data.}
\label{fig:video}
\end{figure}

\begin{table}[t]
\begin{center}
\setlength{\tabcolsep}{2.0mm}{
\begin{tabular}{l|c|c|c|c}
% \hline
\toprule
% \noalign{\smallskip}
& {FID $\downarrow$} & {CSIM $\uparrow$} & {AED $\downarrow$} & {AU-H $\downarrow$} \\
\hline
% \noalign{\smallskip}
{w/o De-exp}  & {13.4} & {0.712} & {0.301} & {0.208}  \\
{w/o MFAT}    & {12.2} & {0.730} & {0.286} & {0.183} \\
{w/o SGMM}           & {13.1} & {0.739} & {0.264} & {0.179}  \\
{w/o Multi-scale}    & {11.7} & {0.734} & {0.272} & {0.180} \\
{w/o FD}             & {18.4} & {0.701} & {0.340} & {0.215} \\
{w/o Comp. D}  & {12.6} & {0.747} & {0.278} & {0.181} \\
% {w/o ID loss}         & {11.5} & {0.676} & {0.260} & {0.177} \\
{Ours}        & \textbf{11.2}  & \textbf{0.769} & \textbf{0.258} & \textbf{0.174}  \\
% \hline
\bottomrule
\end{tabular}}
\end{center}
\caption{Quantitative evaluation of ablation studies.}
\label{table:metrics3}
\end{table}

Our approach is suitable for high-quality expression transfer, as shown in Figure \ref{fig:show}. The source face texture preservation and reference face expression transfer are well achieved. In addition, our method not only directly translates the expression of the reference face image, but also progressively controls the degree of expression change. With the help of our geometry-aware expression representation, we are able to manipulate the face expression change by interpolating the rendered face 3D vertices from the original face expression to the reference face expression. Figure \ref{fig:gradual} shows the effect of this progressive manipulation. During the gradual expression change, the generated face expressions are very natural and do not produce artifacts.

We also try to apply our method to video scenarios. As shown in Figure \ref{fig:video}, unlike the face reenactment methods, the driving result of our method does not have any head posture movement, but only the expression motion will follow the continuous change of the driving video. Self- and Cross-identity driven results are presented.
% and a limitation analysis of the video scenes

\begin{figure}[t]
\begin{center}
\includegraphics[width=1.0\linewidth]{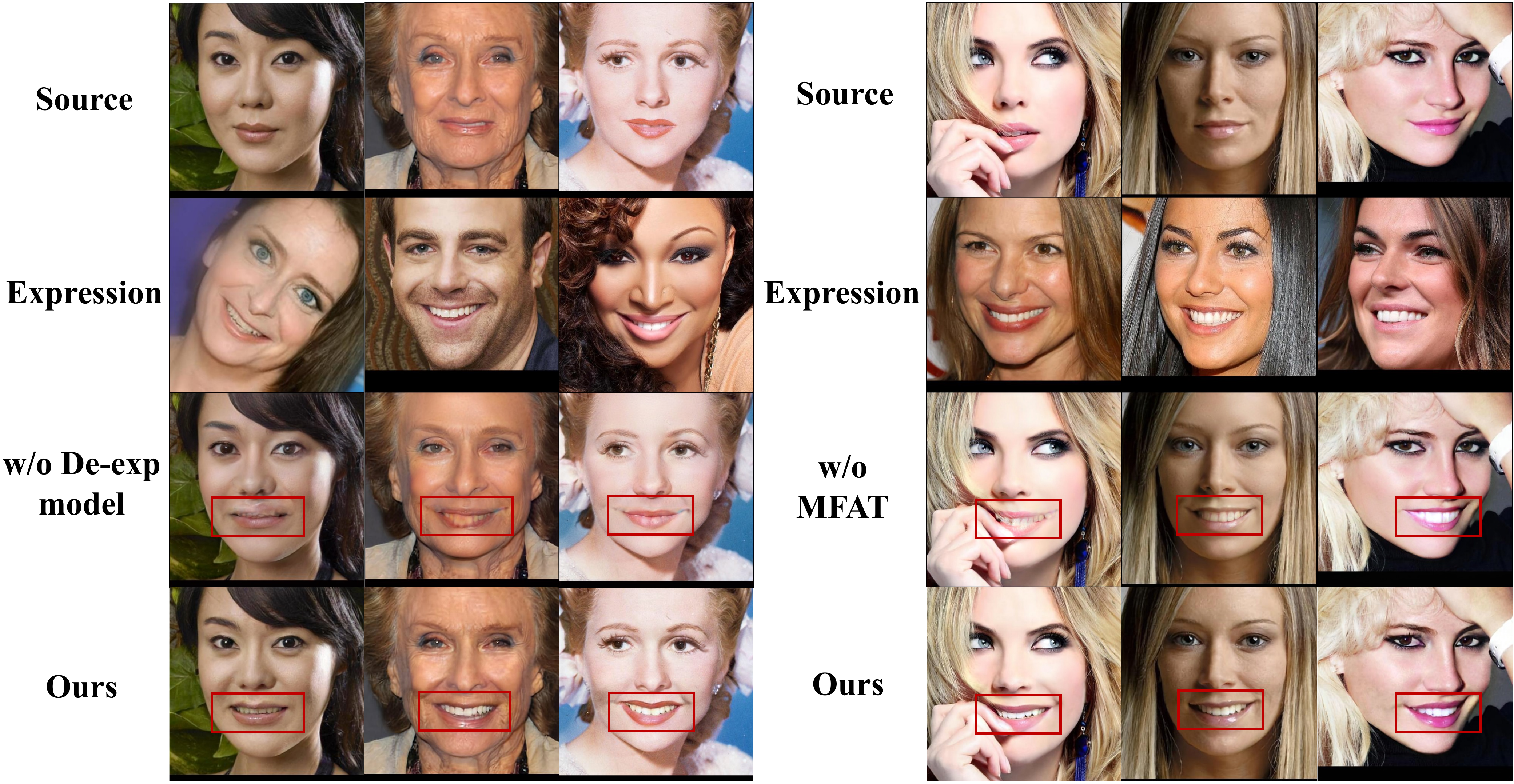}
\end{center}
\caption{Qualitative comparisons of ablation studies on key modules.}
\label{fig:ablation}
\end{figure}

\begin{figure}[t]
\begin{center}
\includegraphics[width=1.0\linewidth]{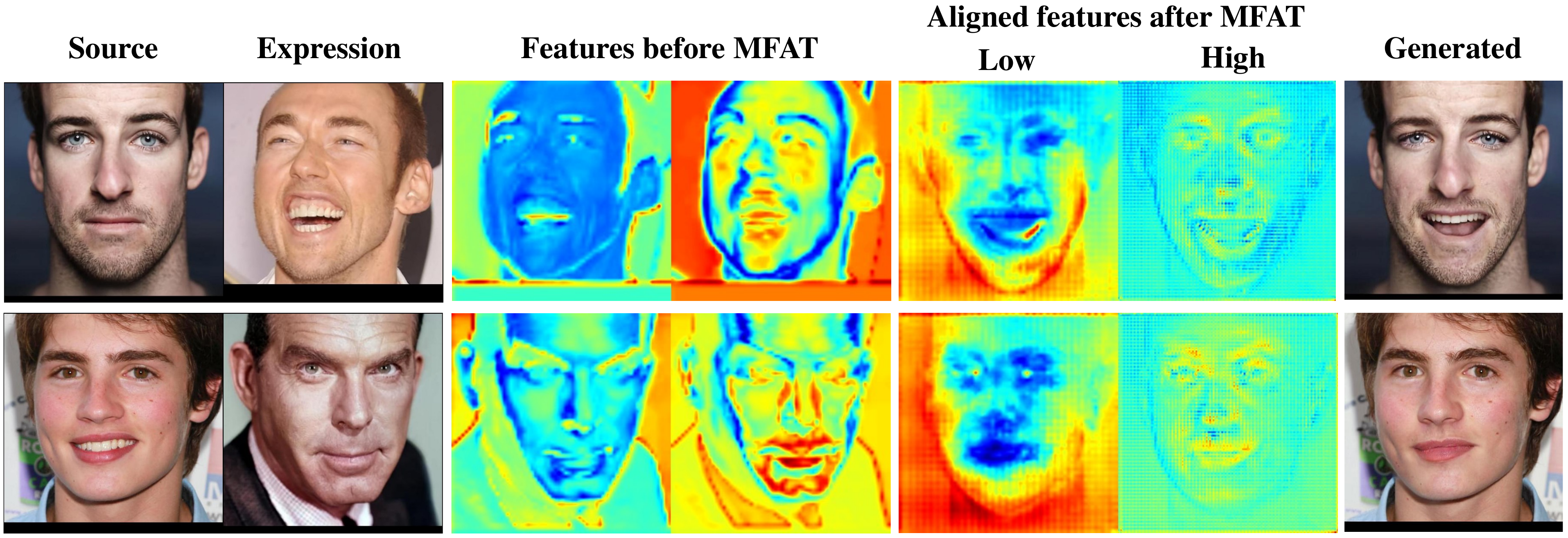}
\end{center}
\caption{Feature visualization before and after the MFAT.}
\label{fig:vis}
\end{figure}

% \begin{table}[h]
% \begin{center}
% \setlength{\tabcolsep}{1.0mm}{
% \begin{tabular}{l|c|c|c|c}
% % \hline
% \toprule
% % \noalign{\smallskip}
% & {FID $\downarrow$} & {CSIM $\uparrow$} & {AED $\downarrow$} & {AU-H $\downarrow$}\\
% \hline
% % \noalign{\smallskip}
% % {each $\lambda$ is 1.0} & {12.0} & {0.723} & {0.919} & {0.834} \\
% {w/o $\mathcal{L}^{\text{pseudo}}_{\text{rec}}$(De-exp)} & {13.4} & {0.712} & {0.301} & {0.208} \\
% {w/o $\mathcal{L}_{\text {exp}}^{\text{shape}}$} & {18.3} & {0.735} & {0.903} & {0.798} \\
% {w/o $\mathcal{L}_{\text{exp}}^{\text{AU}}$} & {11.6} & {0.696} & {0.891} & {0.745} \\
% {w/o $\mathcal{L}_{\text{cx}}$} & {36.5} & {0.764} & {0.996} & {0.882} \\
% {w/o $\mathcal{L}_{\text{adv}}^{\text{global}}$} & {24.6} & {0.735} & {0.951} & {0.813} \\
% {w/o $\mathcal{L}_{\text{adv}}^{\text{comp}}$(Comp. D)} & {12.6} & {0.747} & {0.278} & {0.181} \\
% {Ours} & {\textbf{11.2}} & {\textbf{0.769}} & {\textbf{0.258}} & {\textbf{0.174}} \\
% % \hline
% \bottomrule
% \end{tabular}}
% \end{center}
% \caption{Ablation studies on the losses.}
% \label{table:losses}
% \end{table}

\subsection{Ablation Study}

In this section, we conduct the ablation studies on several important components of our approach. 
% First, for our expression translation method, we use unpaired training data, which leaves us without the ground truth of the faces after the transferred expressions for supervised training. It is difficult for the model to learn accurate color and texture preservation. The pseudo-paired data training method based on our De-expression model constructed in this paper can effectively solve this problem. Furthermore, the Multi-level Feature Aligned Transformer (MFAT) proposed in this paper is an effective method to enhance the details of expressions, which can complement the missing texture details of geometry-aware expression representation, such as teeth part. 
% Table \ref{table:metrics3} shows that there is a significant drop in the quality of the generated images and the accuracy of the expression transfer when these two model parts are not used.
The ablation studies includes: 1.remove the De-expression model; 2. remove the MFAT module 3. remove the global-modulation part of SGMM; 4. Comparison of multi-scale and single-scale attention; 5. Remove feature deformation module; 6. Remove component discriminator. Table \ref{table:metrics3} shows that there is some drop in the quality of the generated images and the accuracy of the expression transfer when removing these modules.
In addition, Figure \ref{fig:ablation} intuitively shows the ablation results of two sets of innovative modules we proposed. The deformation of the mouth and the generation of teeth are the most difficult cases. When the De-expression model is removed, it can be seen from the position marked in the red box that the mouth is not opened normally. Some tooth texture details are missing or incorrect when the MFAT module is not used. In addition, Figure \ref{fig:vis} visualizes the input and output features of the MFAT module. It can be seen that the output features have been aligned with the source face shape, and the high-level features seem to pay attention to details.

% When the De-expression model is not used, from the positions marked by the red boxes in the figure, we can see that the local texture has changed, such as the original beard color has changed and the mouth is not properly closed. When the MFAT module is not used, some local expression details are missing, such as the teeth marked in the figure.
% After using this module, details like teeth will be generated more realistically. 

\section{Conclusion}
In this paper, we propose a geometry-aware facial expression transfer method, coupled with a Multi-level Feature Aligned Transformer and a De-expression model, to achieve excellent expression transfer performance. Unlike previous expression manipulation schemes, we omit video or labeled data support, instead we can learn on more diverse ``in-the-wild" data. Experimental results show that our method can generate high-resolution, high-quality, and accurate facial expression images. The limitations and bad cases are shown in the supplementary material. The future work is to steadily promote it to fit large-angle poses and exaggerated expressions.

% video scenes, and hope to see consistent face expression performance across multiple frames.

\section{Acknowledgments}
This work is supported by the National Key Research and Development Program of China under Grant No. 2021YFC3320103. In addition, we thank Kang Zhao for his help with this paper.

{\small
\bibliographystyle{ieee_fullname}
\bibliography{iccv}
}

\end{document}